\documentclass[runningheads]{llncs}
\usepackage[T1]{fontenc}
\usepackage{graphicx}
\usepackage{hyperref}

\usepackage{multirow}
\usepackage{booktabs}
\usepackage{caption}
\usepackage{siunitx}
\usepackage{perpage} 
\MakePerPage{footnote} 
\begin{document}
\title{Scaling nnU-Net for CBCT Segmentation}
%
%
\author{
Fabian Isensee$^*$\inst{1,2} \and
Yannick Kirchhoff$^*$\inst{2,3,4} \and
Lars Kraemer \inst{1,2} \and
Maximilian Rokuss \inst{2,3} \and
Constantin Ulrich \inst{2,5} \and
Klaus H. Maier-Hein\inst{2,6}
}
\authorrunning{Isensee, Kirchhoff et al.}
%
\institute{
Helmholtz Imaging, German Cancer Research Center (DKFZ), Heidelberg, Germany \and
Division of Medical Image Computing, German Cancer Research Center (DKFZ), Heidelberg, Germany \and
Faculty of Mathematics and Computer Science, Heidelberg University, Germany \and
HIDSS4Health - Helmholtz Information and Data Science School for Health, Karlsruhe/Heidelberg, Germany \and
Medical Faculty Heidelberg, Heidelberg University, Heidelberg, Germany \and
Pattern Analysis and Learning Group, Department of Radiation Oncology, Heidelberg University Hospital
}

\maketitle              
\begin{abstract}
This paper presents our approach to scaling the nnU-Net framework for multi-structure segmentation on Cone Beam Computed Tomography (CBCT) images, specifically in the scope of the ToothFairy2 Challenge. We leveraged the nnU-Net ResEnc L model, introducing key modifications to patch size, network topology, and data augmentation strategies to address the unique challenges of dental CBCT imaging. Our method achieved a mean Dice coefficient of 0.9253 and HD95 of 18.472 on the test set, securing a mean rank of 4.6 and with it the first place in the ToothFairy2 challenge. The source code is publicly available, encouraging further research and development in the field.

\end{abstract}

\def\thefootnote{*}\footnotetext{Equal contribution. Authors are permitted to list their name first in their CVs.}\def\thefootnote{\arabic{footnote}}
\section{Introduction}
Segmenting dental structures such as jaw bones, nerves, and teeth is a critical problem in the field of computer vision due to its significant implications for precision in dental diagnostics, treatment planning, and surgical procedures. Accurate segmentation enables the automated analysis of dental imagery, facilitating tasks such as the identification of dental diseases, the planning of implants, and the navigation of complex anatomical regions during surgery. The high variability in dental anatomy, the proximity of critical structures, and the need for precise localization underscore the importance of developing robust segmentation algorithms. Furthermore, advancements in this area can enhance the efficiency and accuracy of dental care, ultimately leading to improved patient outcomes and personalized treatment strategies.

So far, progress in the field has been held back by a lack of large publicly available datasets. This recently changed with the emergence of the \href{https://toothfairy2.grand-challenge.org/toothfairy2/}{ToothFairy2 Challenge} \cite{lumetti2024enhancing,cipriano2022improving,cipriano2022deep} which provides 480 fully annotated images. Annotations cover 42 classes ranging from bones to alveolar canals, sinuses, implants and individual teeth (according to FDI notation).

This manuscript describes our participation in the \href{https://toothfairy2.grand-challenge.org/toothfairy2/}{ToothFairy2 Challenge}. We base our solution on the recently released nnU-Net ResEnc L \cite{isensee2021nnu,isensee2024nnu} model. It makes use of the standard nnU-Net framework but replaces the canonical U-Net \cite{ronneberger2015u} with a residual encoder U-Net and grants more VRAM budget to the automatic experiment planning procedure.

\section{Methods}
\label{sect:methods}
Our method largely builds on the nnU-Net ResEnc L configuration \cite{isensee2024nnu}. For the sake of clarity we only describe the generated configuration and the changes made to this configuration and the nnU-Net framework as part of our challenge participation. We refer the interested reader to \cite{isensee2021nnu,isensee2024nnu} for more information about nnU-Net. 

\subsection{Baseline configuration}
The generated configuration for the ToothFairy2 dataset has a patch size of 112x224x256, a batch size of 2 and a network topology that encompasses 6 resolution stages. Resampling to a common target spacing is not required since all image already share the same 0.3x0.3x0.3mm spacing. CBCT intensities are normalized with nnU-Net's 'CT' normalization scheme, i.e. each image was clipped to $[-992, 3513]$ followed by subtracting $811$ and dividing by $1001$. 

\subsection{Our improvements}
\subsubsection*{Patch size}
The ToothFairy2 dataset has a median image size of 169x347x371 voxels. Based on past experience we know that large patch sizes enable the network to better learn spatial correspondences and co-localizations, an important prerequisite for correctly classifying all teeth according to the correct FDI notation. Thus, we increase the patch size to 160x320x320. We refrain from increasing it beyond the median image size as this may cause issues with instance normalization layers.

\subsubsection*{Network topology}
The network depth is increased from 6 to 7 resolution stages, resulting in an input stride of (32, 64, 64) in the bottleneck, up from (16, 32, 32) in the baseline configuration. This enables the network to make better use of the increased context information given by the larger patch size. The additional resolution stage follows the same scheme as the network topology generation in nnU-Net ResEnc L and adds an additional 6 residual blocks in the encoder and one convolution in the decoder.

\subsubsection*{Data augmentation}
Correct left/right distinction is pivotal in ToothFairy2. By default, nnU-Net uses mirroring along all spatial axes during training. Usually this is not a cause for concern as anatomical cues like the placement of organs can enable the network to learn the proper left/right assignment even in the presence of mirrored images. In CBCT of the jaw however the available cues seem to be scarce leading to a degradation in performance. To counteract this we disable mirroring augmentation on the left/right axis.

\subsubsection*{Training duration}
Longer training times typically improve segmentation results, with diminishing returns. Given the large dataset size and its complexity we extended the default training duration of 1000 to 1500 epochs.

\subsubsection*{Postprocessing}
The evaluation of the challenge highly rewards true negative predictions by assigning them a HD95 of 0 and a Dice of 1 while strongly punishing false positive/negative predictions \footnote{see \href{https://github.com/AImageLab-zip/ToothFairy}{https://github.com/AImageLab-zip/ToothFairy}}. We exploit this approach by removing small (and likely false positive) predictions by replacing them with background pixels. The cutoffs for discarding predictions are optimized on the training set predictions obtained via a five-fold cross-validation. We optimize the cutoff for each class and evaluation metric (HD95, Dice) independently. For each class, we pick the smaller cutoff between the two metrics as the final cutoff. Due to the fact that the test cases have the same field of view as the 'F' cases and that the 'P' cases suffer from cut-off teeth we optimize the cutoffs on the 'F' cases only.

\subsection{Things we tried that didn't work}
\subsubsection*{Pretraining}
Pretraining with MultiTalent \cite{ulrich2023multitalent} yielded equivalent results as training models from scratch. We experimented with different learning rates and warmup schedules but were unable to improve results.

\subsubsection*{Disabling mirroring entirely}
Our discussion about mirroring above could lead to the hypothesis that mirroring should be entirely disabled for ToothFairy2. However, this leads to reduced segmentation performance. Apparently there are sufficient anatomical cues for the model to make effective use of top/bottom and from/back mirrored images during training. Disabling mirroring entirely might have led to a reduction in training data variability, weakening the model’s ability to generalize.

\subsubsection*{Tooth instance segmentation}
We experimented with a separate network that performs tooth instance segmentation as a first processing step. The idea was to first get the boundaries of individual teeth and only then assign the correct FDI number, thus alleviating problems of teeth being fragmented into two different classes by a standard segmentation approach. The tooth instance segmentation network uses a border-core representation from which tooth instances can be recovered. While this yields improved results relative to our final solution pre-postprocessing it does not improve upon it post-postprocessing (even if we apply postprocessing to both solutions!). We believe this could be related to the '2P' cases which have a reduced field of view and could interfere with the border-core formulation as many teeth are not fully within image boundaries. In addition, the semantic segmentation focused evaluation scheme may favour semantic segmentations-centric approaches.

\section{Results}
Model development is performed by optimizing metrics on a five-fold cross-validation on the ToothFairy2 training split (n=480). The training set consists of two data sources, 'F' and 'P'. The 'P' cases have a reduced field of view with some of the teeth cut off at the image border. As outlined by the challenge organizers on the website, the test cases will follow the 'F' field of view \footnote{\href{https://toothfairy2.grand-challenge.org/faq/}{https://toothfairy2.grand-challenge.org/faq/}}, which is why we put particular emphasis on this subgroup. 

All our models are trained from scratch unless noted otherwise. Model training is done on either a single Nvidia GH200 (96GB) or 2xNvidia A100 (40GB). 

\subsection{Quantitative Results (Training Set)}

Table \ref{tab:performance_metrics} shows a summary of our cross-validation experiments. The nnU-Net ResEnc L baseline delivers solid performance but falls short in the F-cases. Disabling left/right mirroring drastically improves results. Increasing the training duration from 1000 to 1500 epochs has a minor effect but overall improves performance slightly on all cases. Finally, increasing the patch size to 160x320x320 yields the best overall model. 

Among the things that didn't work, disabling mirroring data augmentation entirely yields reduced results over disabling just left/right mirroring. Even though MultiTaltent \cite{ulrich2023multitalent} pretrained models perform better pre-postprocessing, the models trained from scratch surpass them when postprocessing is applied. 

We note that the reported postprocessed metrics may be optimistic because they are reported on the same cases the cutoffs were optimized on. Due to the hidden nature of the test set we were unable to independently verify the generalizability of the cutoff values. Given more time we would have considered using a separate split on the training set for this purpose.

\begin{table}[ht]
\centering
\caption{Five-fold cross-validation results. Table shows performance metrics for different models on F-cases and All cases. Metrics include Dice coefficient and HD95 (Hausdorff Distance at 95th percentile). PP= postprocessed}
\scalebox{0.7}{ 
\begin{tabular}{lcccccccc}
\toprule
\multirow{2}{*}{Model} & \multicolumn{4}{c}{F-cases (n=63)} & \multicolumn{4}{c}{All cases (n=480)} \\
\cmidrule(lr){2-5} \cmidrule(lr){6-9}
& \multicolumn{1}{c}{Dice} & \multicolumn{1}{c}{HD95} & \multicolumn{1}{c}{Dice (PP)} & \multicolumn{1}{c}{HD95 (PP)} & \multicolumn{1}{c}{Dice} & \multicolumn{1}{c}{HD95} & \multicolumn{1}{c}{Dice (PP)} & \multicolumn{1}{c}{HD95 (PP)} \\
\midrule
nnU-Net ResEncL default & 0.7445 & 24.99 & 0.7808 & 20.23 & 0.8211 & 22.135 & 0.8948 & 13.682 \\
 $+$ no l/r mirr & 0.9083 & 9.758 & 0.9302 & 6.212 & 0.8610 & 17.417 & 0.9210 & 11.091 \\
 $+$ no l/r mirr, 1500ep & 0.9064 & 10.354 & 0.9291 & 6.262 & 0.8624 & 17.392 & 0.9212 & 11.038 \\
 $+$ no mirr, 1500ep, ps160x320x320 & 0.8938 & 12.62 & 0.9255 & 7.266 & 0.8622 & 17.531 & 0.9223 & 11.239 \\
 $+$ no l/r mirr, 1500ep, ps160x320x320, pretr & \textbf{0.9119} & \textbf{9.258} & 0.9320 & \textbf{5.875} & \textbf{0.8691} & \textbf{16.454} & 0.9242 & 11.001 \\
 $+$ no l/r mirr, 1500ep, ps160x320x320 & 0.9098 & 10.14 & \textbf{0.9344} & 6.069 & 0.8686 & 16.726 & \textbf{0.9271} & \textbf{10.605} \\
\bottomrule
\end{tabular}
} 
\label{tab:performance_metrics}
\end{table}

\subsection{Qualitative Results (Training Set)}

\begin{figure}[t!]
    \centering
    \includegraphics[width=\textwidth]{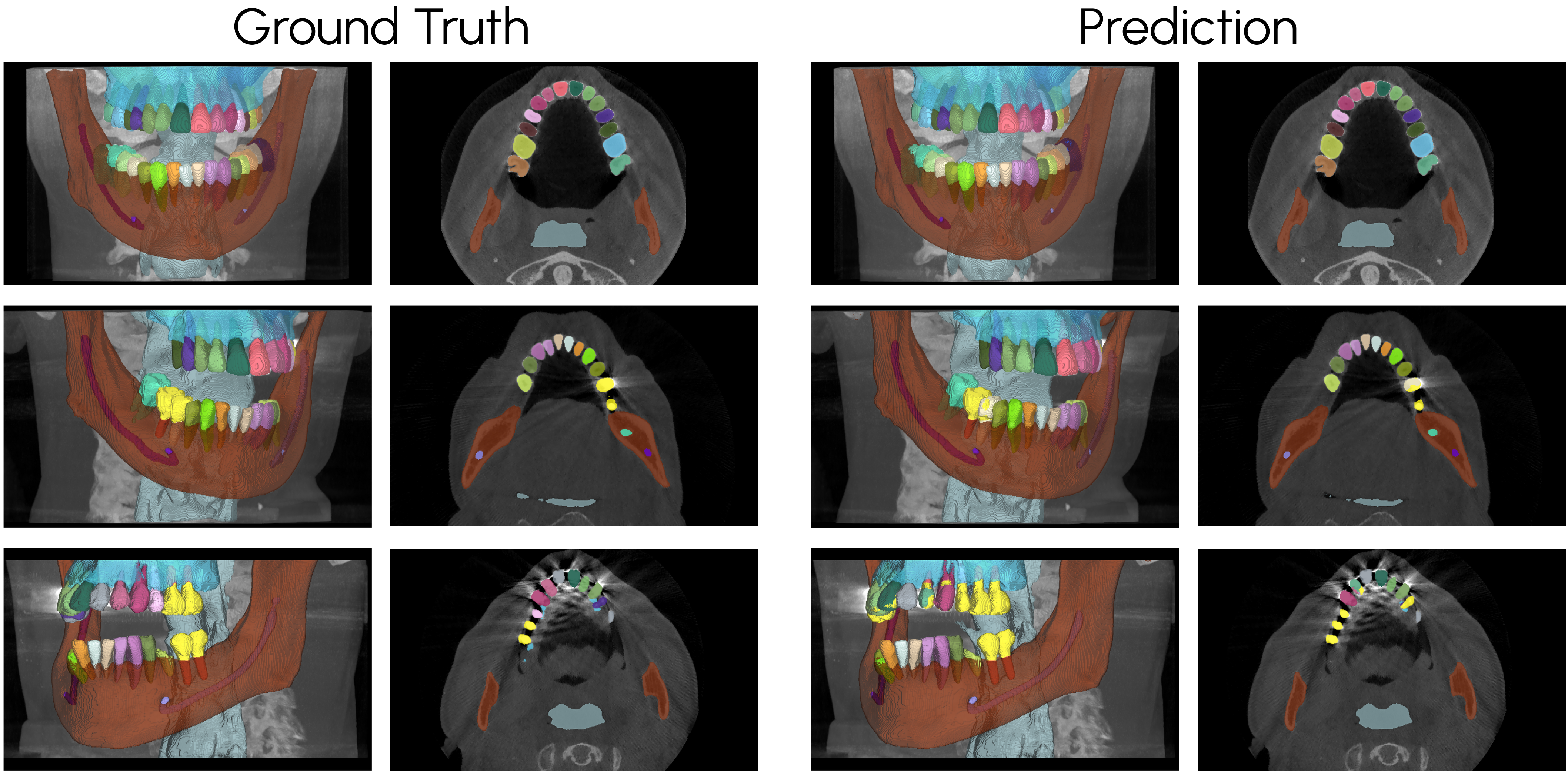}
    \caption{Qualitative Results of our method (single model), predictions are obtained from the validation sets of each fold in our 5-fold cross-validation. Ground Truth is shown on the left with both, a 3D rendering and a representative 2D slice. Correspondingly, the predictions are shown on the right. Most cases are predicted accurately and all teeth are correctly classified (top). Small errors include misclassification of teeth or inconsistencies between crowns and teeth (middle). Severe artifacts can cause poor performance (bottom).}
    \label{fig:qualitative_results}    
\end{figure}

Figure \ref{fig:qualitative_results} provides a visual comparison between our method's predictions and the ground truth, selected based on model performance in the cross-validation on the training set (good, medium, bad). In the majority of cases, our method delivers reliable and precise results, underscoring its effectiveness in accurately segmenting and classifying complex structures. This demonstrates the model's capability to generalize across diverse anatomical variations and imaging conditions.
Typical failure modes include teeth being split into multiple classes or teeth being incorrectly classified. These errors are often attributed to the presence of dental crowns, which introduce significant variations in appearance and can confound the model's performance. 

\begin{figure}[t!]
    \centering
    \includegraphics[width=\textwidth]{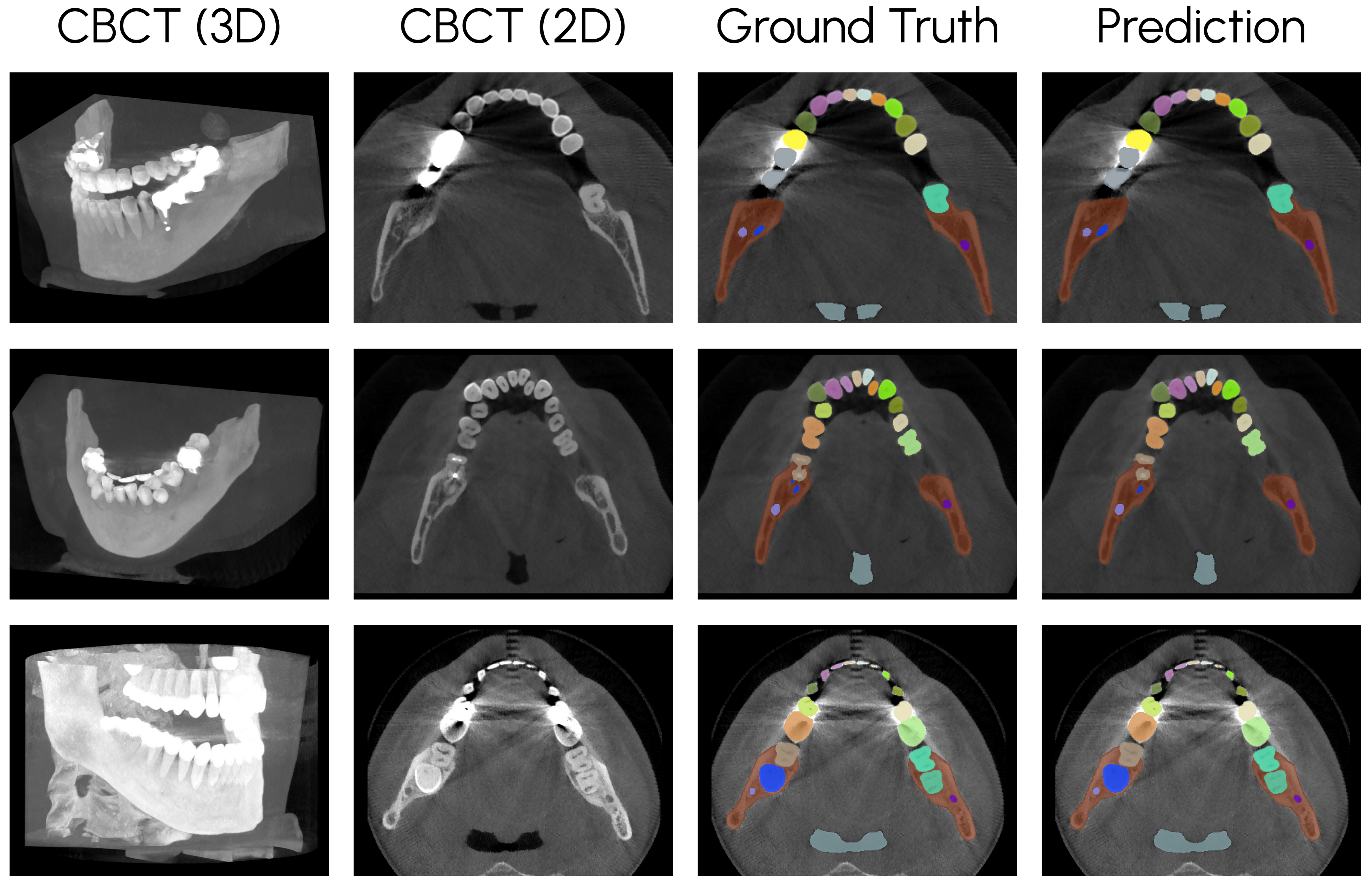}
    \caption{Illustration of our method's robust performance on challenging cases, including artifacts, noise, and anatomical outliers. Again the predictions are obtained from the validation sets of our 5-fold cross-validation.}
    \label{fig:qualitative_cases}    
\end{figure}

Additionally, we identified particularly challenging images from the training set by searching for imaging artifacts and unusual tooth placements. As shown in Figure \ref{fig:qualitative_cases} our method is highly robust with respect to such conditions and produces accurate results even in cases where it is difficult to discern tooth from background by the human eye. Likewise, anatomical diversity within the dataset is handled robustly and our model is able to still assign the correct FDI notation even when teeth are displaced.

\subsection{Test Set Submission}
For our final submission we train two models (as described in Section \ref{sect:methods}) from scratch on all available training data (n=480). Non-deterministic training (no seeding) ensures some diversity in the models outputs that we leverage by ensembling them for our final prediction. Postprocessing is applied using the cutoffs determined on the five-fold cross-validation. 

\subsection{Quantitative Results (Test Set)}

\begin{table}[ht]
\centering
\caption{Performance on the Test Set (n=50)}
\begin{tabular}{lccc}
\toprule
\textbf{Model} & \textbf{Dice} & \textbf{HD95} & \textbf{Mean Rank} \\
\midrule
Ours  & 0.9253 & 18.472 & 4.6 \\
\bottomrule
\end{tabular}
\label{tab:test_set_results}
\end{table}


Table \ref{tab:test_set_results} summarizes our test set performance. We observe a slight drop off in model performance relative to the training set cross-validation. The observed drop in performance on the test set is likely due to domain shift, which can arise from differences in imaging protocols or patient demographics between the training and test data. To what extend the test set images differ from the training set is not disclosed by the challenge organizers.

The ranking scheme of the challenge computes mean Dice and HD95 for all classes and then ranks all participants for all classes and metrics (42 classes x 2 metrics = 84 ranks). The mean rank is then used to determine the final ranking of the challenge. Our model won the ToothFairy2 challenge with a mean rank of 4.6, narrowly edging out the users 'Oculins' (mean rank 4.8) and 'Mors' (mean rank 5.2)\footnote{Leaderboard: \href{https://toothfairy2.grand-challenge.org/evaluation/final-test-phase-phase/leaderboard/}{https://toothfairy2.grand-challenge.org/evaluation/final-test-phase-phase/leaderboard/}}.

\section{Conclusion}
In this work, we explored the adaptation and enhancement of the nnU-Net ResEnc L model for the segmentation of CBCT images within the context of the ToothFairy2 Challenge. By carefully modifying key parameters such as patch size, network topology, and data augmentation strategies, we significantly improved the model's performance on this challenging dataset. Our approach effectively addresses the unique challenges posed by dental CBCT images, including the need for precise localization and class distinction, which are critical for accurate dental diagnostics and treatment planning.

Our experimental results demonstrate that our tailored nnU-Net configuration, combined with strategic postprocessing techniques, can achieve high levels of accuracy as evidenced by the Dice coefficient and HD95 metrics across both the F-cases and the entire dataset. 

Although some advanced techniques like pretraining with MultiTalent and tooth instance segmentation did not improve results, our final model configuration achieved the first place on the leaderboard, highlighting its robustness and effectiveness.

Source code is publicly available as part of the nnU-Net repository: \\ \href{https://github.com/MIC-DKFZ/nnUNet/tree/master/documentation/competitions/Toothfairy2}
{\texttt{https://github.com/MIC-DKFZ/nnUNet/tree/master/documentation/\\competitions/Toothfairy2}}.

\begin{credits}
\subsubsection{\ackname}
Part of this work was funded by Helmholtz Imaging (HI), a platform of the Helmholtz Incubator on Information and Data Science. The present contribution is supported by the Helmholtz Association under the joint research school "HIDSS4Health – Helmholtz Information and Data Science School for Health".
\end{credits}

\bibliographystyle{abbrv}
\bibliography{bib}

\end{document}